\documentclass[preprint,11pt]{elsarticle}

\usepackage[margin=2.4cm]{geometry}
\RequirePackage[colorlinks]{hyperref}
\RequirePackage{amsthm,amsmath,amssymb,bbm}
\RequirePackage[]{natbib}

 \usepackage{rotating}
\usepackage{colortbl}
\usepackage{multirow}
\usepackage[table]{xcolor}
\usepackage{makecell}

\newcommand{\R}{\mathbb{R}}

\newcommand{\A}{\mathbf{A}}

\numberwithin{equation}{section}
\theoremstyle{plain}
\newtheorem{theorem}{Theorem}
\newtheorem{assume}{Assumption}
\newtheorem{lemma}{Lemma}

\newtheorem{rmk}{Remark}

\allowdisplaybreaks

\begin{document}
\begin{frontmatter}
\title{Optimal sparse phase retrieval via a quasi-Bayesian approach}

		\author[aaatienmt]{The Tien Mai}\ead{the.tien.mai@fhi.no}
		
		\affiliation[aaatienmt]{
			organization={
				Norwegian Institute of Public Health},
			city={Oslo},
			postcode={0456}, 
			country={Norway}}

\begin{abstract}
This paper addresses the problem of sparse phase retrieval, a fundamental inverse problem in applied mathematics, physics, and engineering, where a signal need to be reconstructed using only the magnitude of its transformation while phase information remains inaccessible. Leveraging the inherent sparsity of many real-world signals, we introduce a novel sparse quasi-Bayesian approach and provide the first theoretical guarantees for such an approach. Specifically, we employ a scaled Student’s t-distribution as a continuous shrinkage prior to enforce sparsity and analyze the method using the PAC-Bayesian inequality framework. Our results establish that the proposed Bayesian estimator achieves minimax-optimal convergence rates under sub-exponential noise, matching those of state-of-the-art frequentist methods. To ensure computational feasibility, we develop an efficient Langevin Monte Carlo sampling algorithm. Through numerical experiments, we demonstrate that our method performs comparably to existing frequentist techniques, highlighting its potential as a principled alternative for sparse phase retrieval in noisy settings.
\end{abstract}
		

\end{frontmatter}		


\section{Introduction}
\label{sc_intro}

The phase retrieval problem is a fundamental inverse problem in applied mathematics, physics, and engineering. The objective is to reconstruct a signal—such as a function, image, or wave—using only the magnitude of its transformation, without access to phase information \cite{fienup1987phase, millane1990phase,candes2013phaselift,candes2014solving,shechtman2015phase, jaganathan2016phase}. In many practical scenarios, direct phase measurements are either infeasible or impossible due to inherent physical constraints. As a result, phase retrieval techniques are essential in applications where only intensity measurements are available, such as optical imaging, crystallography, and astronomical imaging.  

In recent years, phase retrieval has attracted growing interest in the statistical community \cite{lecue2015minimax, cai2016optimal, wang2017sparse, yuan2019phase, sun2018geometric, wu2023nearly}. While theoretical results establish that an arbitrary signal can be reconstructed, many applications involve signals that exhibit natural sparsity \cite{jaganathan2016phase}. Exploiting this sparsity allows for signal recovery using significantly fewer measurements, improving both efficiency and feasibility. Consequently, the development of robust algorithms that incorporate sparsity constraints has become a key focus, leading to advances in optimization, regularization techniques, and computational complexity.  

In this paper, we consider the following problem of recovering a signal vector \( \theta^* \in \mathbb{R}^p \) from a set of quadratic measurements:  
\begin{equation}
\label{eq_main_model}
y_j = (a_j^\top \theta^*)^2 + \varepsilon_j, \quad j = 1, \ldots, m,
\end{equation}  
where \( \varepsilon_j \) is a stochastic noise term with zero mean and $ a_j $ are sensing vectors. The goal is to identify a point that is close either $ \theta^* $ or $ -\theta^* $. We refer interested readers to \cite{shechtman2015phase} and references therein for a detailed discussion of the scientific and engineering motivations behind this model. In many applications, particularly in imaging, the signal \( \theta^* \) admits a sparse representation under a known deterministic linear transformation. Without loss of generality, we assume that such a transformation has already been applied, so that \( \theta^* \) itself is sparse. Under this assumption, model \eqref{eq_main_model} corresponds to the sparse phase retrieval problem.  

Various approaches have been developed to address sparse phase retrieval, incorporating different strategies to enforce sparsity. A common technique is explicit regularization through penalty terms \cite{CCG15, OYDS12} or direct sparsity constraints \cite{JH19, NJS15, SBE14}. Alternatively, some methods promote sparsity within algorithmic frameworks, such as thresholding procedures \cite{WZGAC18, ZWGC18}. Empirical risk minimization with sparsity constraints has also been investigated \cite{LM15}.  

Several notable algorithms have been proposed in this domain. Thresholded Wirtinger flow \cite{candes2015phase, cai2016optimal} provides an efficient iterative method for phase retrieval. In the context of non-convex optimization, \cite{wu2023nearly} introduced a mirror descent approach incorporating sparsity constraints to handle noisy measurements. Approximate Message Passing, which leverages sparsity-inducing priors, was explored in \cite{SR15} under a different modeling framework. Additionally, a variational approach was studied in \cite{zhang2022phase}, while maximum a posteriori estimation frameworks were considered in \cite{dremeau2015phase, dremeau2017phase}.

Despite existing efforts to incorporate Bayesian methods into sparse phase retrieval, no  theoretical justification has been established for such approaches. In this paper, we introduce a novel quasi-Bayesian framework for sparse phase retrieval and provide a rigorous theoretical foundation for its application. Moreover, we demonstrate that our Bayesian approach achieves a convergence rate comparable to that of frequentist methods \cite{lecue2015minimax, cai2016optimal, wu2023nearly}.  

We establish convergence rates for our proposed method in the setting of noisy sparse phase retrieval under sub-exponential noise. Notably, our rate is minimax optimal, consistent with existing results in the frequentist literature \cite{lecue2015minimax, cai2016optimal, wu2023nearly}. To the best of our knowledge, this represents the first theoretical guarantee of its kind for a Bayesian approach in sparse phase retrieval. 
Our theoretical analysis is based on the PAC-Bayesian inequality framework, originally introduced to establish generalization error bounds for learning algorithms \cite{McA, STW} (here, PAC stands for Probably Approximately Correct). Later, \cite{catonibook} demonstrated that this technique can also be used to derive oracle inequalities and excess risk bounds, an approach we adopt in our theoretical development. For a detailed introduction and recent advances in this area, we refer to \cite{alquier2024user} and \cite{guedj2019primer}.

To incorporate sparsity, we employ a continuous shrinkage prior. Various continuous shrinkage priors have been explored in the literature to promote sparsity \cite{li2017variable, piironen2017sparsity, bhadra2019lasso}. In particular, we use a scaled Student’s t-distribution as our prior, a choice that has been successfully applied to other sparse estimation problems \cite{dalalyan2012mirror, dalalyan2012sparse, mai2023high, mai2024sparse}. To facilitate computation, we propose an efficient approximate sampling method based on Langevin Monte Carlo, ensuring the tractability and scalability of our Bayesian framework.  

Finally, we conduct numerical experiments to evaluate the performance of our proposed method in comparison with the approach presented in \cite{wu2023nearly}. The results indicate that our method achieves comparable performance, demonstrating its effectiveness in sparse phase retrieval.  

The rest of the paper is presented as follow. In Section \ref{sc_problem_method}, we formally present the problem of sparse phase retrieval and introduce our quasi-Bayesian approach for this problem. in Section \ref{sc_theory}, we present non-assymptotic results showing that our approach can recover the underlying signal at the minimax-optimal rate. We also provide contraction rate of the quasi-posterior in this section. In section \ref{sc_numberical}, we present our implemnetation and numerical studies. Short conclustion are given in Section \ref{sc_conlsution}. All technical proofs are given in  \ref{sc_proofs}.

\section{Problem and method}
\label{sc_problem_method}

\subsection{Sparse phase retrieval}

The objective of sparse phase retrieval is to recover an unknown sparse signal vector \( \theta^* \in \mathbb{R}^p \), where the sparsity of the signal is characterized by \( s^* := \|\theta^*\|_0 \), denoting the count of non-zero components. This reconstruction is performed based on a set of possibly noisy quadratic measurements:
\begin{equation}
\label{eq_main}
y_j = (\mathbf{A}_j^\top\theta^*)^2 + \varepsilon_j, \quad j = 1, \dots, m.
\end{equation}  
Here, we adopt the standard setting in which the sensing vectors \( (\mathbf{A}_j)_{j=1}^m \) are independently drawn from a Gaussian distribution, i.e., \( \mathbf{A} \sim \mathcal{N} (0,\mathbf{I}_p) \). Additionally, the noise terms \( \varepsilon_j \) are independent, centered, and sub-exponential. This noise modeling assumption is particularly relevant in practical applications of phase retrieval—especially in optics—where heavy-tailed noise can emerge due to photon counting effects \cite{shechtman2015phase,cai2016optimal}. 

For clarity in our theoretical analysis, we focus on the case of real-valued signals and real Gaussian measurement vectors in the context of noisy sparse phase retrieval. More specifically, it is assumed that $ s^* < p $.

Given the observations \( \{ \mathbf{A}_j, y_j \}_{j=1}^m \), several established methods \cite{cai2016optimal, WZGAC18, YWW19, ZWGC18} have focused on optimization-based approaches using the following empirical risk function:  

\begin{equation}
\label{eq:objective_function}
r(\theta) = \frac{1}{4m} \sum_{j=1}^m \big((\mathbf{A}_j^\top\theta)^2 - y_j\big)^2
.
\end{equation}  
Alternative risk functions based on amplitude measurements \( |\mathbf{A}_j^\top\theta^*| \) has also been explored in previous studies \cite{WZGAC18, ZWGC18}. However, such an approach leads to a non-smooth objective function, which poses additional challenges for optimization. 

Instead of relying on a point estimation framework, this work adopts a probabilistic approach. Notably, our methodology exhibits a connection to the Bayesian framework in a specific case, which will be elaborated on in a subsequent section. Due to this connection, we refer to our method as a quasi-Bayesian approach.

\subsection{A quasi-Bayesian approach}

Rather than consider an empirical risk minimization approach to infer the signal of interest, we consider a distribution (posterior) based the exponential of the negative empirical risk. For any \(\lambda > 0\),  we consider the following Gibbs posterior \(\hat{\rho}_{\lambda} \),
\begin{equation}
\label{eq_Gibbs_poste}
\hat{\rho}_{\lambda} (\theta) 
\propto 
\exp[-\lambda r (\theta)] \pi(\theta)
, 	
\end{equation}
where $ \pi $ is a prior distribution  given in \eqref{eq_priordsitrbution} that promotes sparsity. 

The purpose of the Gibbs posterior in \eqref{eq_Gibbs_poste} is to adjust the distribution towards parameter values that minimize the empirical risk on the sample, with the tuning parameter $\lambda$ determining the extent of this adjustment, which will be further discussed in upcoming sections. Specifically, if $ \pi $ assigns higher probability to sparse vectors, $\hat{\rho}_\lambda $ will favor sparse vectors with low empirical risk, satisfying our requirements. A more detailed discussion on this topic can be found in \cite{catonibook, guedj2019primer, alquier2024user}.

The reason that the Gibbs posterior, defined in \eqref{eq_Gibbs_poste}, is referred to as a quasi-Bayesian method is that: When the noise variable \( \varepsilon_j \) follows a Gaussian distribution \( \mathcal{N} (0, \sigma^2) \), setting \( \lambda = 2m/\sigma^2 \) transforms our Gibbs posterior into a standard Bayesian posterior. However, our framework extends beyond this classical setting by accommodating a broader class of noise distributions, specifically sub-exponential noise, which is more appropriate for the phase retrieval problem \cite{shechtman2015phase,cai2016optimal}.

In this study, we employ the following prior distribution, proposed in \cite{dalalyan2012mirror,dalalyan2012sparse}.
For a fixed constant \( H_1 >0 \), for all $ \theta \in \mathbb{R}^p  $ that \(  \|\theta\|_2 \leq H_1 \), we adopt the scaled Student distribution as our prior distribution.
\begin{eqnarray}
\label{eq_priordsitrbution}
\pi (\theta) 
\propto 
\prod_{j=1}^{p} 
(\varsigma^2 + \theta_{i}^2)^{-2}
,
\end{eqnarray}
where $ \varsigma>0 $ is a tuning parameter.   In this framework, $ H_1 $ functions as a regularization parameter, typically assumed to be very large. Consequently, the distribution of \( \pi \) is approximately that of \( S\varsigma \sqrt{2} \), where \( S \) is a random vector with components independently drawn from a Student's t-distribution with 3 degrees of freedom. By choosing a very small value for \( \varsigma \), most entries of \( \varsigma S \) are concentrated near zero. However, the heavy-tailed nature of the distribution ensures that a few components deviate significantly from zero. This property effectively encourages sparsity in the parameter vector through the prior.

\section{Theoretical guarantees}
\label{sc_theory}

In this section, we present the fundamental theoretical results underpinning our proposed method.

\subsection{Assumptions}

To establish our theoretical results, we begin by specifying the key conditions underpinning the analysis.

\begin{assume}
	\label{assume_X_bounded}
Assume that there exists a constant $ C >0 $ such that for all $ u \in \mathbb{R}^p $,
	$ | \A_j^\top u | \leq C < \infty $.
\end{assume}

\begin{assume}
	\label{assum_heavy_tailed}
Assume that the noise variables  $ \varepsilon_1, \ldots, \varepsilon_m $ are independent 
and there exist two known
	constants $\sigma>0$ and $\xi>0$ such that
	$$  \mathbb{E} (| \varepsilon_j |^{k}) \leq 
	\sigma^{2} \frac{k!}{2} \xi^{k-2},
	 \quad \forall k\geq 3
	.
	$$
\end{assume}

\begin{assume}
	\label{assume_mendelson}
	Assume that there exists a constant $ \kappa_0 >0 $ such that for every $ s, t \in \mathbb{R}^p $,
	$$ 
	\mathbb{E} | (\A^\top s ) (\A^\top t ) | 
	\geq  
	\kappa_0 \| s \| \| t \| 
	.
	$$
\end{assume}

Assumption \ref{assume_X_bounded} is a reasonable condition ensuring that the model in \eqref{eq_main} has a well-defined finite mean, i.e., \( \mathbb{E}(y_j) < \infty \).  Assumption \ref{assum_heavy_tailed} states that the noise term \( \varepsilon_j \) follows a sub-exponential distribution, a standard assumption in the analysis of machine learning algorithms \cite{boucheron2013concentration}. It is worth noting that this noise condition is practically motivated and aligns with assumptions commonly adopted in the literature; see, for example, \cite{shechtman2015phase, lecue2015minimax, cai2016optimal, wu2023nearly}. This encompasses cases such as Gaussian noise \( \mathcal{N}(0, \sigma^2) \) (where \( \xi = \sigma \)) and any centered random variable that is absolutely bounded by some constant \( C > 0 \) (where \( \sigma = \xi = C \)), as elaborated in Chapter 2 of \cite{boucheron2013concentration}.  Assumption \ref{assume_mendelson} is fairly mild and is satisfied by many common choices of random vectors. For further discussion, we refer to \cite{lecue2015minimax} and \cite{eldar2014phase}.

\subsection{Main results}

Due to the inherent structure of the measurements in \eqref{eq_main}, it is impossible to distinguish between \( \theta^* \) and \( -\theta^* \). Consequently, the most achievable outcome is to construct an estimator \( \hat{\theta} \) that closely approximates one of these two solutions \cite{lecue2015minimax,cai2016optimal,wu2023nearly}. The primary objective of this work is to develop a method that achieves this goal while analyzing how the Euclidean distance between \( \hat{\theta} \) and either \( \theta^* \) or \( -\theta^* \) is influenced by factors such as sparsity, dimensionality, and the number of measurements.

Put $ C_1 = 8 ( \sigma^2 + C^2 ), C_2 = 2^6 \max(\xi,C) C $ and 
$$
\lambda^* = m/(C_1 +2C_2),
\quad
\varsigma^* = ( 4Cpm )^{-1} .
$$ 

\begin{theorem}
	\label{thm_heavy_tailed_loss}
	Assume that Assumption \ref{assume_X_bounded}, \ref{assum_heavy_tailed} and \ref{assume_mendelson} are satisfied. Take $\lambda = \lambda^* $, $ \varsigma = \varsigma^* $.  Then for all $ \theta^* $ such that $  \| \theta^*\|_2 \leq H_1 - 2p \varsigma $, we have with probability at least $ 1-\delta, \delta\in (0,1) $ that
	\begin{equation*}
	\mathbb{E}_{\theta\sim\hat{\rho}_{\lambda}}
	\left[ \| \theta - \theta^* \|^2 \| \theta + \theta^*  \|^2  \right]
	\leq 
	\mathfrak{C}\,
	\sigma^2 \frac{s^* \log ( m p / s^*)  +	\log( 2 / \delta)
	}{ m }
	,
	\end{equation*}	
	for some universal constant $ \mathfrak{C}  > 0 $ depending only on $ H_1, C,\kappa_0, \xi $.
\end{theorem}

\begin{rmk}
From Theorem \ref{thm_heavy_tailed_loss}, we demonstrate that our proposed method achieves a fast error rate of order $ \sigma^2 s^* \log(p/s^*)/m $. the results show an adaptive nature that does not require the knowledge of the sparsity level $ s^* $. 
This rate is known to be minimax-optimal	as a lower bound of the same order was established in \cite{lecue2015minimax}. Similar results were also obtain in \cite{cai2016optimal} and \cite{wu2023nearly}. However, our results are pioneering, as they present the first of their kind for a Bayesian method applied to this problem.
\end{rmk}

The proofs are presented in Appendix \ref{sc_proofs}, where we adopt the `PAC-Bayesian bounds' methodology introduced in \cite{catonibook} as the central technical framework. For a more comprehensive treatment of PAC-Bayes theory and its recent developments, we refer the reader to \cite{guedj2019primer, alquier2024user}. PAC-Bayesian bounds offer a theoretical foundation for assessing the generalization error of learning algorithms by integrating Bayesian principles into the Probably Approximately Correct (PAC) learning paradigm. In contrast to traditional PAC bounds, which primarily focus on point estimators, PAC-Bayes bounds consider a distribution over hypotheses, effectively balancing empirical error and model complexity via the Kullback–Leibler divergence between the posterior and prior distributions \cite{STW, McA}. Moreover, as highlighted in \cite{catoni2004statistical, catonibook}, this approach is particularly valuable for deriving non-asymptotic guarantees on learning performance. PAC-Bayesian bounds have found widespread applications across diverse domains, including sparse estimation and classification \cite{dalalyan2008aggregation, alquier2011PAC, guedj2013pac, luu2019pac, mai2023high}, as well as matrix estimation problems \cite{mai2015, mai2023bilinear, mai2023reduced, mai2025misclassification}.

In addition to the result established in Theorem \ref{thm_heavy_tailed_loss}, we further derive a complementary asymptotic property concerning the behavior of the quasi-posterior distribution. Specifically, we obtain a result that is commonly referred to in the Bayesian literature as the contraction rate of the posterior. This notion characterizes the speed at which the quasi-posterior concentrates around the true parameter value as the sample size increases, thereby providing a theoretical guarantee on the efficiency of the learning procedure under the specified modeling assumptions.

\begin{theorem}
	\label{thrm_contraction_slow} 
Consider the same set of assumptions as those stated in Theorem~\ref{thm_heavy_tailed_loss}, along with the previously defined quantities \( \varsigma \) and \( \lambda \). Let \( \delta_m \in (0, 1) \) denote a sequence that converges to zero as the sample size \( m \to \infty \).  Define
	\begin{align*}
	\Theta_m 	
	= 
	\Biggl\{ \theta \in \mathbb{R}^{p}: 
 \| \theta - \theta^* \|^2 \| \theta + \theta^*  \|^2  
\leq 
\mathfrak{C}\,
\sigma^2 \frac{s^* \log ( m p / s^*)  +	\log( 2 / \delta_m)
}{ m }
	\Biggl\}
	.
	\end{align*}
	Then
	$$ \mathbb{E} \Bigl[ \mathbb{P}_{ \hat{\theta} \sim \hat{\rho}_{\lambda}} 
	( 	\hat{\theta} \in \Theta_m ) \Bigr] 
	\geq 
	1- 2\delta_m \xrightarrow[n\rightarrow \infty]{} 1.  
	$$
\end{theorem}

In this section, we presented the selected values for the tuning parameters \( \lambda \) and \( \varsigma \) in our proposed method, which provide different theoretical prediction error rates. Although these values serve as a useful starting point, they may not be optimal in practical applications. Users can, for instance, apply cross-validation to fine-tune these parameters for specific tasks. Still, the theoretical values discussed here offer a valuable reference for assessing the appropriate scale of tuning parameters in real-world contexts.

\section{Numerical studies}
\label{sc_numberical}
\subsection{Implementation using a gradient-based sampling method}
In this study, we utilize the Langevin Monte Carlo (LMC) method as a sampling technique for exploring the quasi-posterior distribution. LMC is particularly well-suited for this task, as it leverages gradient information to guide the sampling process efficiently, improving exploration of the target distribution, especially in high-dimensional settings \cite{durmus2017nonasymptotic,dalalyan2017theoretical,durmus2019high}. A crucial aspect of the method involves computing the gradient of the logarithm of the quasi-posterior, which plays a key role in determining the dynamics of the Langevin diffusion. In the case of our quasi-posterior in \eqref{eq_Gibbs_poste}, this gradient can be explicitly expressed as follows:
\begin{align*}
\nabla \log \hat{\rho}_{\lambda} (\theta) 
\propto
-  \frac{\lambda}{m} \sum_{j=1}^m [(\mathbf{A}_j^\top\theta)^2 - y_j ](\mathbf{A}_j^\top\theta)\mathbf{A}_j  -4 \sum_{\ell =1}^{p} \theta_{\ell}/(\varsigma^2 + \theta_{\ell}^2).
\end{align*}
Our methodology relies on the constant step-size unadjusted LMC algorithm \cite{durmus2019high}, which begins with an initial point \( \theta_0 \) and updates iteratively according to the following rule:
\begin{equation}
\label{langevinMC}
\theta_{k+1} = \theta_{k} - \gamma \nabla \log \hat{\rho}_{\lambda}(\theta_k) + \sqrt{2\gamma} N_k \qquad k=0,1,\ldots
\end{equation}
In this formulation, \( \gamma > 0 \) represents the step-size, while \( N_0, N_1, \ldots \) denote independent random vectors whose components follow an i.i.d. standard Gaussian distribution. The choice of \( \gamma \) is critical, as an overly small step-size can introduce instability in the summation, a challenge discussed in \cite{roberts2002langevin}.

In addition to LMC, we also investigate the Metropolis-adjusted Langevin algorithm (MALA), which enhances the sampling process by integrating a Metropolis–Hastings (MH) correction. However, this correction introduces an acceptance/rejection step at each iteration, potentially leading to slower performance. Following the recommendations in \cite{roberts1998optimal}, the step-size \( \gamma \) in MALA is chosen to achieve an acceptance rate of approximately 0.5. Under identical conditions, LMC is employed with a smaller step-size than MALA to balance efficiency and stability.

Our R code is available at: \url{https://github.com/tienmt/sparse_Bayes_phase_retrieval}.

\subsection{Simulation setup}

We consider a similar manner to simulate data as in \cite{wu2023nearly}. More specifically, in all experiments, we generate a $s^* $-sparse vector $ \theta^* \in \R^p $ by sampling $\theta^*_i $ i.i.d from $ \mathcal{N} (0,1) $ for $i=1,\dots,p $ and setting $(p-s^*)$ random entries of $ \theta^* $ to zero. Then $ \theta^* $ is rescaled such that $ \|\theta^*\|_2 = 1 $ so that we can easily access the effect of noise-to-signal ratio $ \sigma/ \|\theta^*\|_2 $. We sample $m$ i.i.d.\ measurement vectors $\A_j\sim \mathcal{N} (0, \mathbf{I}_p)$ and noise terms $\varepsilon_j\sim \mathcal{N} (0, \sigma^2)$, and then  the observations are generated, as in \eqref{eq_main}, $ y_j = (\A_j^\top \theta^*)^2 + \varepsilon_j, j =1,\ldots, m $.

To further evaluate and compare the performance of our methods, we benchmark LMC and MALA against the mirror descent approach, recently proposed and analyzed in \cite{wu2023nearly}. This method is referred to as `MD’ in the following, and its implementation is publicly available alongside the original publication. For our experiments, LMC and MALA are each run for 30,000 iterations, with the first 1,000 iterations discarded as a burn-in period. The MD method is used with its default settings, running for 5,000 iterations. Our theoretical insights from Section \ref{sc_theory} suggest that setting \( \varsigma < 1 \) encourages sparsity, while the regularization parameter \( \lambda \) should scale with the sample size \( m \). Guided by these considerations, we fix \( \varsigma = 0.1 \) and \( \lambda = 4m \) in all experiments.

In this section, we present numerical experiments to assess the performance of our proposed methods in comparison with the MD method. Specifically, we investigate how the minimum relative estimation error, defined as  
\[
\mathrm{mre} := \frac{\min \{ \| \theta - \theta^* \|^2 , \| \theta + \theta^* \|^2 \} }{ p \| \theta^* \|^2},
\]  
varies with the noise-to-signal ratio \( \sigma / \|\theta^*\|_2 \), the sample size \( m \) and the sparsity level \( s^* \). The results are conducted over 100 independent Monte Carlo simulations.

\subsection{Simulation results}

\subsubsection{Sample size effect:} 
We begin by investigating how the performance of the methods varies with the sample size \( m \). To isolate the effect of \( m \), we fix the dimensionality at \( p = 100 \), the true sparsity level at \( s^* = 10 \), and the noise level at \( \sigma = 1 \). The sample sizes considered are \( m = 100, 200, 500, 1000, 2000 \), covering a range from low to moderately large data regimes. For each value of \( m \), we generate 100 independent datasets and compute the relative estimation errors for LMC, MALA, and MD.

\begin{figure}[!h]
\centering
\includegraphics[width=10cm,height=5cm]{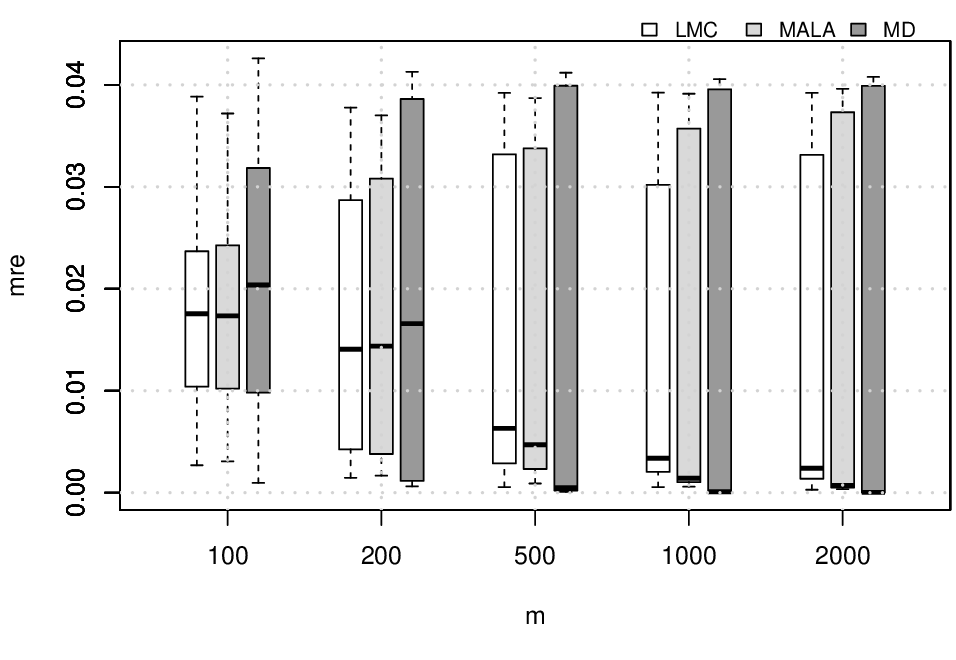}
\caption{The relation between the minimum relative error and the sample size.}
	\label{fg_sample_siez}
\end{figure} 

Figure \ref{fg_sample_siez} presents the distribution of errors across simulations in the form of boxplots. The results indicate that, although some variability remains, the median and lower quantiles of the minimum relative estimation error steadily improve as the sample size increases. This trend is consistent across all three methods, with LMC and MALA demonstrating slightly more stable performance in larger sample settings. 
It is also observed that in scenarios with smaller sample sizes, specifically \( m = 100 \) and \( m = 200 \), our proposed methods—LMC and MALA—tend to yield slightly lower estimation errors compared to the MD approach. However, as the sample size increases, particularly at \( m = 2000 \), these differences become marginal and practically negligible. This suggests that the advantage of LMC and MALA is more pronounced in low-data regimes, while all methods perform comparably when ample data is available.

The decreasing error with growing \( m \) aligns with our theoretical expectations and supports the validity of our assumptions.

\subsubsection{Noise-to-signal effect:} 
Next, we examine the impact of the noise-to-signal ratio \( \sigma / \| \theta^* \|_2 \), on the performance of the methods. Since we set \( \| \theta^* \|_2 = 1 \), this ratio simplifies to the noise level itself. To isolate this effect, we fix the dimensionality at \( p = 100 \), the sample size at \( m = 500 \), and the true sparsity level at \( s^* = 10 \). The noise levels considered are \( \sigma = 0.5, 1, 2, 5, 10 \), covering both low- and high-noise regimes. For each value of \( \sigma \), we generate 100 independent datasets and compute the relative estimation errors for the LMC, MALA, and MD methods.

\begin{figure}[!h]
\centering
\includegraphics[width=10cm,height=5cm]{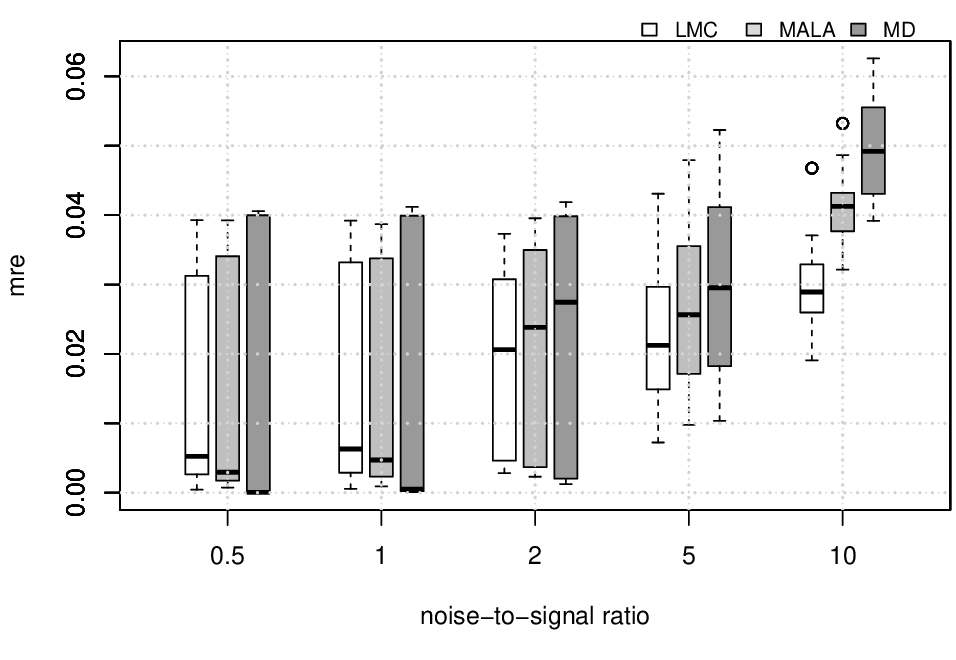}	
\caption{The relation between the minimum relative error and the noise-to-signal-ratio.}
	\label{fg_noise_signal}
\end{figure}

The results, summarized in the boxplots in Figure \ref{fg_noise_signal}, reveal a clear dependence of the minimum relative error on the noise level, in agreement with our theoretical findings. When the noise-to-signal ratio is moderate (\( \sigma / \| \theta^* \|_2 \leq 1 \)), the MD method exhibits slightly lower estimation error on average, although with noticeably higher variability. In contrast, in higher noise settings (\( \sigma / \| \theta^* \|_2 > 1 \)), both LMC and MALA outperform MD, with LMC achieving the most favorable results in terms of accuracy and stability. These observations highlight the robustness of our methods under increased noise conditions.

\subsubsection{Sparsity effect:} 

We now investigate the influence of the sparsity level \( s^* \) on the performance of the methods. To this end, we fix the sample size at \( m = 1000 \), the dimensionality at \( p = 500 \), and the noise level at \( \sigma = 1 \), while varying the sparsity level as \( s^* = 5, 20, 100, 250, 500 \). For each value of \( s^* \), we generate 100 independent datasets and compute the relative estimation errors for the LMC, MALA, and MD methods.

\begin{figure}[!h]
\centering
\includegraphics[width=10cm,height=5cm]{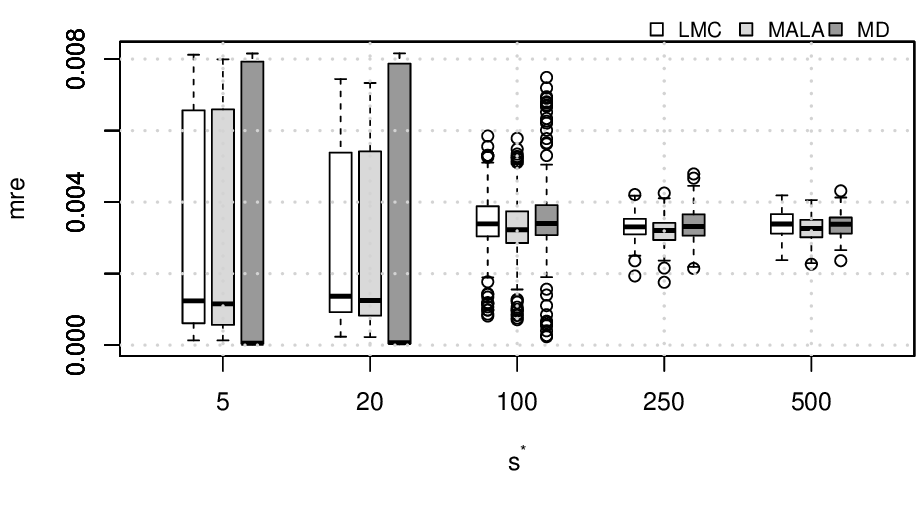}	
\caption{The relation between the minimum relative error and the sparsity.}
	\label{fg_sparsity}
\end{figure} 

The results, summarized in the boxplots in Figure \ref{fg_sparsity}, indicate that estimation error increases as the level of sparsity decreases—that is, as \( s^* \) becomes larger. When the underlying signal is highly sparse (i.e., small \( s^* \)), all three methods achieve low estimation error on average, though with noticeably greater variability across simulations. In contrast, as the sparsity decreases (i.e., \( s^* \) increases), the performance of all methods converges, and their behavior becomes more stable and consistent. These findings highlight the sensitivity of all methods to the degree of sparsity, particularly in regimes of strong signal compression.

\subsubsection{Tuning parameter effect:}
\begin{figure}[!h]
\centering
\includegraphics[width=7.5cm,height=4.5cm]{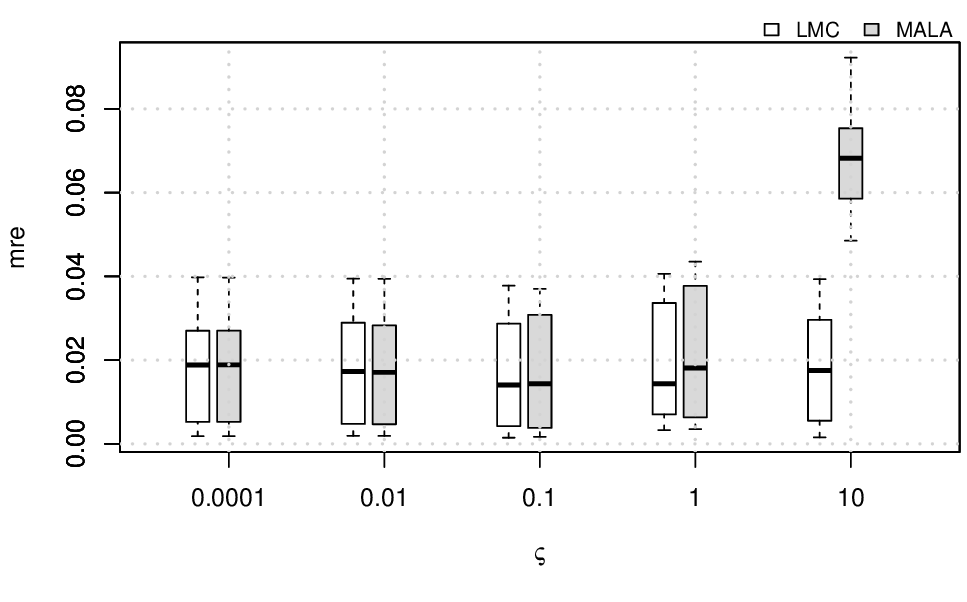}
\includegraphics[width=7.5cm,height=4.5cm]{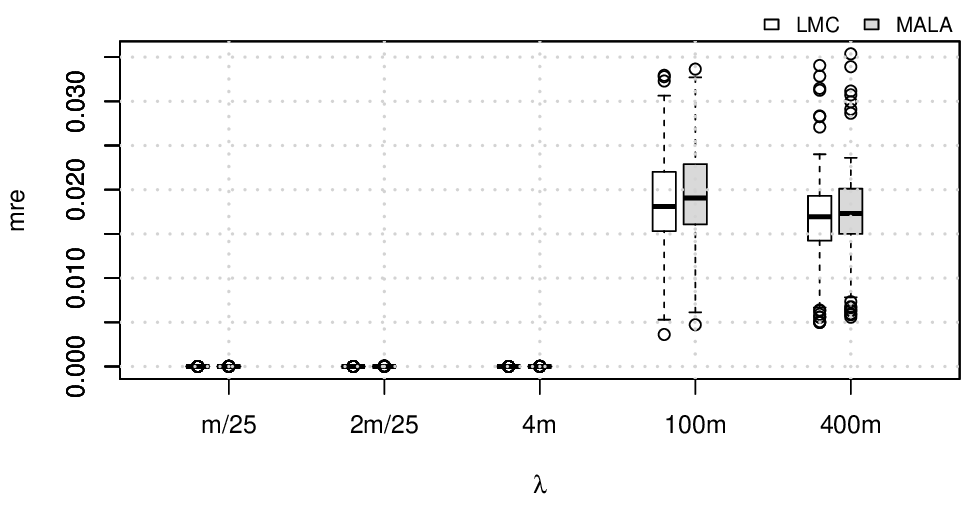}
\caption{The relation between the minimum relative error and the tuning parameters. Left: changing $ \varsigma $, Right: changing $ \lambda $.}
	\label{fg_change_varsigma}
\end{figure}

We now examine the impact of the parameter \( \varsigma \) in the prior. We set \( m = 200 \), \( p = 100 \), \( s^* = 10 \), and \( \sigma = 1 \), and vary \( \varsigma \) across the values \( 0.0001, 0.01, 0.1, 1, 10 \). The results from 100 simulations are presented in Figure \ref{fg_change_varsigma}. The figure clearly illustrates that \( \varsigma < 1 \) are good choices, while values of \( \varsigma > 1 \) lead to poor performance. The selected value \( \varsigma = 0.1 \) yields the smallest error.

To evaluate the sensitivity of the parameter \( \lambda \), we fix \( m = 50 \), \( p = 100 \), \( s^* = 10 \), \( \sigma = 1 \), and \( \tau = 0.1 \), and vary \( \lambda \) across the values \( m/25, 2m/25, 4m, 100m, 400m \). As shown in Figure \ref{fg_change_varsigma}, setting \( \lambda \) too large can severely degrade the performance of the algorithm. While other values that smaller than $ 4m $ seems to work similar.

\subsection{Application: Reconstruction of Handwritten Digit Images}

To illustrate the effectiveness of our proposed methodology, we consider the task of reconstructing handwritten digit images using data from the \texttt{MNIST}  dataset \cite{lecun1998gradient}, a widely recognized benchmark in the fields of machine learning and image analysis. The dataset consists of grayscale images of handwritten digits ranging from 0 to 9, standardized in format and resolution.

For the purpose of this analysis, we randomly select representative samples of digit 2 and digit 4, which are treated as the ground-truth signal vectors, denoted by \( \theta^* \). Each image is represented as a two-dimensional array, with digit 2 being a \( 16 \times 22 \) pixel image and digit 4 a \( 20 \times 22 \) pixel image. These images exhibit inherent sparsity, quantified by the proportion of non-zero-valued pixels: 42.6\% for digit 2 and 34.1\% for digit 4, respectively.

\begin{figure}[!ht]
\centering
\includegraphics[width=9cm,height=2.5cm]{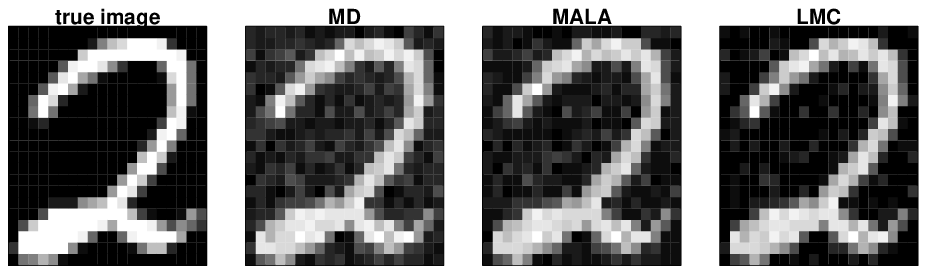}
\\
\includegraphics[width=9cm,height=2.5cm]{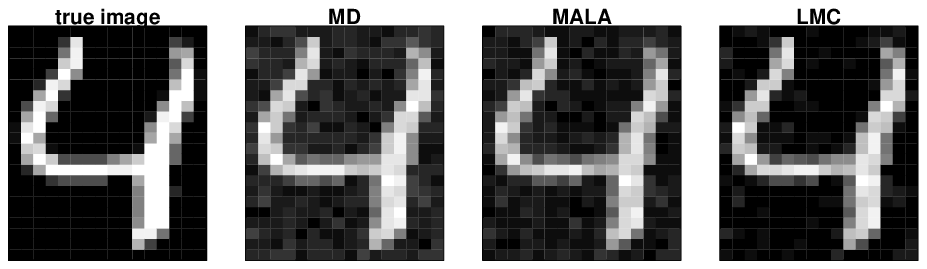}	
	\caption{Handwritten Digit Images recovered by different methods.}
	\label{fg_realdata}
\end{figure}

To ensure consistency with the assumptions underlying our model and to facilitate the subsequent analysis, we normalize the pixel intensity values so that $ \|\theta^*\|_2 = 1 $. The normalized images are shown in Figure~\ref{fg_realdata}. For the purpose of generating synthetic measurements, we sample \( m = 4000 \) independent and identically distributed measurement vectors \( \A_j \sim \mathcal{N}(0, \mathbf{I}_p) \) and noise terms \( \varepsilon_j \sim \mathcal{N}(0, 1) \). The corresponding observations are then generated according to the model \( y_j = (\A_j^\top \theta^*)^2 + \varepsilon_j \).

The images displayed in Figure~\ref{fg_realdata} demonstrate that all the methods under consideration are capable of recovering the underlying image.

\section{Conclusion}
\label{sc_conlsution}

In this work, we introduced a novel quasi-Bayesian framework for the problem of noisy sparse phase retrieval. We established non-asymptotic upper bounds on the estimation error, demonstrating that our method attains the minimax-optimal rate. Furthermore, we derived the contraction rate of the quasi-posterior, providing a rigorous characterization of its frequentist behavior. These theoretical results are on par with the best-known guarantees in the frequentist literature and represent a significant advancement in the Bayesian treatment of the sparse phase retrieval problem.

To enable practical implementation, we developed efficient sampling algorithms that leverage the gradient structure of the quasi-posterior, specifically using Langevin Monte Carlo (LMC) and its variants. Our numerical experiments, benchmarked against a recent state-of-the-art mirror descent method, show that the proposed quasi-Bayesian procedures are highly competitive in terms of estimation accuracy and robustness, particularly in challenging regimes with limited data or high noise.

Several promising directions remain open for future research. One avenue is to extend the quasi-Bayesian framework to handle more general structured sparsity or hierarchical network structure, which may further improve practical performance. Another important question is the development of adaptive or data-driven choices for hyper-parameters such as using empirical Bayes or using hierarchical priors, which currently rely on theoretical guidance.  Finally, applying the methodology to real-world tasks, such as in optics or signal processing, would further validate its practical utility and reveal new modeling challenges.

\subsubsection*{Acknowledgments}
The views, findings, and opinions presented in this work are exclusively those of the author and do not reflect the official stance of the Norwegian Institute of Public Health.

\subsubsection*{Conflicts of interest/Competing interests}
The author declares no potential conflict of interests.

\appendix
\section{Proofs}
\label{sc_proofs}

Let \( \mathcal{P}(\Theta) \) represent the collection of all probability measures defined on \(\Theta\). We denote the expected risk as
\begin{equation*}
R (\theta) 
= 
\mathbb{E} \big((\A_j^\top\theta)^2 - Y_j\big)^2
,
\end{equation*}
and note that from Pythagorean's theorem we have that
\begin{equation*}
R (\theta) - R (\theta^*) 
= 
\mathbb{E} \left[ (\A_j^\top\theta)^2 - (\A_j^\top\theta^*)^2  \right]^2
.
\end{equation*}

\begin{proof}[\bf Proof of Theorem \ref{thm_heavy_tailed_loss}]
	
The application of Jensen's inequality to inequality \eqref{lemma:exponential:1} in Lemma \ref{lemma:exponential} results in:
	\[
	\mathbb{E} \exp\Biggl[
	\alpha
	\left( \int R {\rm d} \hat{\rho}_{\lambda} - R(\theta^*) \right)
	-
	\lambda\left( \int r {\rm d} \hat{\rho}_{\lambda} - r(\theta^*) \right)
	- 
	\mathcal{K}(\hat{\rho}_{\lambda}, \pi)
	- 
	\log\frac{2}{\delta}\Biggr] 
	\leq
	\frac{\delta}{2}
	.
	\]
	Now, using inequality $ e^x \geq
	\mathbf{1}_{\mathbb{R}_{+}}(x)$, one obtains that
	\begin{equation*}
	\mathbb{P}\Biggl\{ \int R {\rm d} \hat{\rho}_{\lambda} - R(\theta^*)
	\leq
	\frac{ \int r {\rm d} \hat{\rho}_{\lambda} - r(\theta^*) +
		\frac{1}{\lambda}\left[\mathcal{K}(\hat{\rho}_{\lambda}, \pi)
		+ \log\frac{2}{\delta}\right] } {\frac{\alpha}{\lambda} }
	\Biggr\}
	\geq
	1-\frac{\delta}{2}.
	\end{equation*}
	Using Donsker and Varadhan's variational inequality, Lemma \ref{lemma:dv},
	we get
	%
	\begin{equation}\label{interm3bis}
	\mathbb{P}\Biggl\{ \int R {\rm d} \hat{\rho}_{\lambda} - R(\theta^*)
	\leq
\inf_{\rho\in \mathcal{P}(\Theta)}
 \frac{ \int r
		{\rm d}\rho- r(\theta^*) +
		\frac{1}{\lambda}\left[\mathcal{K}(\rho, \pi)
		+ \log\frac{2}{\delta}\right] } {\frac{\alpha}{\lambda} } \Biggr\}
	\geq1-\frac{\delta}{2}
	.
	\end{equation}

	Now, using inequality $ e^x \geq
\mathbf{1}_{\mathbb{R}_{+}}(x)$, from \eqref{lemma:exponential:2} one obtains that
	\begin{equation}
	\label{interm4} \mathbb{P}\Biggl\{ \int r{\rm d}\rho- r(\theta^*)
	\leq\frac{\beta}{\lambda} \left[\int
	R{\rm d}\rho- R(\theta^*) \right] + \frac{1}{\lambda}\left[
	\mathcal{K}(\rho, \pi) + \log\frac{2}{\delta} \right]
	\Biggr\}\geq1 - \frac{\delta}{2}.
	\end{equation}
By applying a union bound argument to (\ref{interm4}) and (\ref{interm3bis}), we derive the following general PAC-Bayesian bound
	%
	\begin{equation}
	\label{PAC_bound_heavytailed}
	\mathbb{P}\Biggl\{\!  \int R {\rm d} \hat{\rho}_{\lambda} - R(\theta^*)
	\leq
\inf_{\rho\in \mathcal{P}(\Theta) } 
\frac{ \beta[\int R{\rm d}\rho-
		R(\theta^*) ] + 2 [
		\mathcal{K}(\rho, \pi) + \log\frac{2}{\delta} ] } {
		\alpha }\! \Biggr\}
	\geq
	1-\delta.
	\end{equation}

We now proceed to derive an explicit error bound from the general PAC-Bayesian bound. The approach involves restricting the infimum on the right-hand side to a specific distribution and computing the bound explicitly.
	We limit the infimumn in \eqref{PAC_bound_heavytailed} to $ \rho:= \pi_0 $ defined in \eqref{eq_specific_distribution}.

One has that
\begin{align*}
R (\theta) - R (\theta^*) 
& =
\mathbb{E} \left[ (\A_j^\top\theta)^2 - (\A_j^\top\theta^*)^2  \right]^2
\\
& =
\mathbb{E} \left[ (\A_j^\top\theta -\A_j^\top\theta^*) (\A_j^\top\theta^* + \A_j^\top\theta)  \right]^2
\\
& \leq
\mathbb{E} \left[ \A_j^\top(\theta - \theta^*) 2C  \right]^2
\\
& \leq
4C^2 
\mathbb{E} \left[ \| \A_j \|^2 \| \theta - \theta^*\|^2 \right]
\\
& \leq
4C^2 
p \| \theta - \theta^*\|^2 
,
\end{align*}	
where we have used that $ | \A_j^\top \theta | \leq C $ for all $ \theta $ and that $ \mathbb{E} \| \A_j \|^2 =p  $ with $ \A_j \sim \mathcal{N} (0, \mathbf{I}_p ) $. Thus, we have,  from Lemma \ref{lema_bound_prior_arnak}, that
\begin{align*}
\int [ R (\theta) - R (\theta^*) ] {\rm d} \pi_0
 \leq
4C^2 p
\int \| \theta - \theta^*\|^2  {\rm d} \pi_0
 \leq
16 C^2 p^2 \varsigma^2 
,
\end{align*}
and
	$
\mathcal{K}(\pi_0 , \pi)
\leq
4 s^* \log \left(\frac{H_1 }{\varsigma s^*}\right)
+
\log(2)
.
$
Thus, we incorporate these bounds into inequality \eqref{PAC_bound_heavytailed}, resulting in:	
	\begin{equation*}
\mathbb{P}\Biggl\{\!  \int R {\rm d} \hat{\rho}_{\lambda} - R(\theta^*)
\leq
	\inf_{\varsigma \in (0, H_1/2p )} 
\!
 \frac{ \beta 16 C^2 p^2 \varsigma^2 
 	+ 
 	2 [ 	4 s^* \log \left(\frac{H_1 }{\varsigma s^*}\right)
 	+
 	\log(2) + \log\frac{2}{\delta} ] } {
	\alpha }
\Biggr\}
\geq
1-\delta
.
\end{equation*}
The choice \( \lambda = \frac{m}{C_1 + 2C_2} \) guarantees that \( \alpha > 0 \) and \( \lambda < \frac{m}{w} \). Under this specification, it follows that \( \beta / \alpha \leq 3 \), and moreover, \( 1/\alpha = \frac{m}{2C_1 + 2C_2} \).
\\
 The choice $ \varsigma = ( 4Cp m )^{-1} $ leads to the conclusion that
	\begin{equation*}
\mathbb{P}\Biggl\{\!  \int R {\rm d} \hat{\rho}_{\lambda} - R(\theta^*)
\leq
\frac{3}{m^2}
	+ 
	4 (C_1 + C_2) \frac{ 4 s^* \log \left(\frac{ H_1 4 C p m  }{ s^*}\right)
	+	\log(2) + \log\frac{2}{\delta}}{ m }
\Biggr\}
\geq
1-\delta
.
\end{equation*}	
Therefore, with probability at least $ 1-\delta $, we obtain from Assumption \ref{assume_mendelson} that 
	\begin{equation*}
  \int \kappa_0 \| \theta - \theta^* \|^2 \| \theta + \theta^*  \|^2  {\rm d} \hat{\rho}_{\lambda} 
\leq
\frac{3}{m^2}
+ 
4(C_1 + C_2)  \frac{ 4 s^* \log \left(\frac{ H_1 4 C p m  }{ s^*}\right)
	+	\log(2) + \log\frac{2}{\delta}}{ m }
.
\end{equation*}	
Thus the results of Theorem \ref{thm_heavy_tailed_loss} is obtained. The proof is completed.

\end{proof}

\begin{proof}[\textbf{Proof of Theorem \ref{thrm_contraction_slow}}]
	Consider any sequence \( \delta_m\in (0,1) \) satisfying \( \delta_m\to 0 \) as \( n \to \infty \). According to inequality \eqref{lemma:exponential:1} stated in Lemma \ref{lemma:exponential}, we obtain the following result:
	\begin{align*}
	\mathbb{E} \int \exp  \Biggl\{  \alpha    \Bigl( R(\theta) - R(\theta^*) \Bigr)
	- \lambda ( r(\theta) - r(\theta^*)  )    
	- \log \left[\frac{{\rm d} \hat{\rho}_{\lambda}}{d \pi} (\theta)  \right]
	- \log\frac{2}{\delta_m}
	\Biggr\}
	\hat{\rho}_{\lambda}(d \theta )
	\leq \frac{\delta_m}{2}
	.
	\end{align*}
	We now apply Chernoff’s bounding technique, which relies on the inequality \( e^x \geq \mathbf{1}_{\mathbb{R}_{+}}(x) \). This leads to the following bound:
	$$
	\mathbb{E} \Bigl[ 
	\mathbb{P}_{\theta \sim \hat{\rho}_{\lambda}} 
	(\theta \in \Theta_m) \Bigr]
	\geq 
	1- \frac{\delta_m}{2}
	$$
	where
	$$
	\mathcal{A}_m
	= 
	\left\{\theta : 	
	\alpha	[R( \theta )-R^* ]
	\leq      
	\lambda ( r(\theta) - r(\theta^*)  )    
	+ 
	\log \left[\frac{{\rm d} \hat{\rho}_{\lambda}}{d \pi} (\theta)  \right]
	+
	\log\frac{2}{\delta_m}
	\right\}.
	$$
By applying Lemma \ref{lemma:dv} together with the definition of $\hat{\rho}_\lambda$, we obtain the following for any $ \theta \in \mathcal{A}_m$:
	\begin{align*}
	\alpha    \Bigl( R(\theta) - R(\theta^*) \Bigr)
	&    \leq  
	\lambda\Bigl( r(\theta) - r(\theta^*) \Bigr)  +       \log \left[\frac{d\hat{\rho}_{\lambda}}{d \pi} (\theta)  \right]
	+ \log\frac{2}{\delta_m}
	\\
	& \leq -\log\int\exp\left[-\lambda r(\theta)\right]\pi({\rm d} \theta) - \lambda r(\theta^*)
	+ \log\frac{2}{\delta_m}
	\\
	& = \lambda\Bigl( \int r(\theta) \hat{\rho}_{\lambda}({\rm d}\theta) - r(\theta^*) \Bigr)  +    \mathcal{K}(\hat{\rho}_\lambda,\pi)
	+ \log\frac{2}{\delta_m}
	\\
	& = \inf_{\rho} \left\{ \lambda\Bigl( \int r(\theta) \rho({\rm d}\theta) - r(\theta^*) \Bigr)  +    \mathcal{K}(\rho,\pi)
	+ \log\frac{2}{\delta_m} \right\}.
	\end{align*}
	
	\noindent Now, let us define
	$$ \mathcal{B}_m= \left\{\forall\rho\text{, }\beta \left(-\int Rd\rho + R(\theta^*) \right)
	+ \lambda \left( \int r d\rho - r(\theta^*) \right) \leq
	\mathcal{K}(\rho, \pi) + \log \frac{2}{\delta_m}\right\}. 
	$$
	
	\noindent Using~\eqref{lemma:exponential:2}, we have that
	$$
	\mathbb{E} \Bigl[\mathbf{1}_{\mathcal{B}_m} \Bigr]
	\geq 
	1 - \frac{\delta_m}{2}
	.
	$$
	We will now prove that, if $\lambda$ is such that $\alpha>0$,
	\begin{equation}
	\label{eq_1_contraction}
	\mathbb{E} \Bigl[ 
	\mathbb{P}_{\theta \sim \hat{\rho}_{\lambda}} (\theta \in \Theta_m) 
	\Bigr] 
	\geq 
	\mathbb{E} \Bigl[ \mathbb{P}_{\theta \sim \hat{\rho}_{\lambda}} (\theta \in\mathcal{A}_m)\mathbf{1}_{\mathcal{B}_m} \Bigr]
	.
	\end{equation}
	In order to do so, assume that we are on the set $\mathcal{B}_m$, and let $ \theta \in\mathcal{A}_m$. Then,
	\begin{align*}
	\alpha    \Bigl( R(\theta) - R(\theta^*) \Bigr)
	& \leq \inf_{\rho} \left\{ \lambda\Bigl( \int r(\theta) \rho({\rm d}\theta) - r(\theta^*) \Bigr)  +    \mathcal{K}(\rho,\pi)
	+ \log\frac{2}{\delta_m} \right\}
	\\
	& \leq \inf_{\rho} \left\{ \beta \Bigl( \int R(\theta) \rho({\rm d}\theta) - R(\theta^*) \Bigr)  +   2 \mathcal{K}(\rho,\pi)
	+ 2 \log\frac{2}{\delta_m} \right\}
	\end{align*}
	that is,
	$$
	R(\theta) - R(\theta^*) 
	\leq 
	\inf_{\rho  } \frac{ \beta \left[\int R d\rho -
		R(\theta^*) \right] + 2 \left[
		\mathcal{K}(\rho, \pi) + \log \frac{2}{\delta_m} \right] } {
		\alpha  }
	.
	$$
	We upper-bound the right-hand side exactly as in the proof of Theorem~\ref{thm_heavy_tailed_loss}, in page \pageref{PAC_bound_heavytailed}, to obtain that
	\begin{equation*}
	\kappa_0 \| \theta - \theta^* \|^2 \| \theta + \theta^*  \|^2 
	\leq
	\frac{3}{m^2}
	+ 
	4(C_1 + C_2)  \frac{ 4 s^* \log \left(\frac{ H_1 4 C pm  }{ s^*}\right)
		+	\log(2) + \log\frac{2}{\delta_m}}{ m }
	.
	\end{equation*}	
	This yields
	$ \theta \in \Theta_m$ and we obtain \eqref{eq_1_contraction}.
	
	Moreover, we have that
	\begin{align*}
	\mathbb{E} \Bigl[ \mathbb{P}_{\theta \sim \hat{\rho}_{\lambda}} (\theta \in\mathcal{A}_m)\mathbf{1}_{\mathcal{B}_m} \Bigr]
	& = 
	\mathbb{E} \Bigl[ (1-\mathbb{P}_{\theta\sim \hat{\rho}_{\lambda}} (\theta \notin\mathcal{A}_m)) (1-\mathbf{1}_{\mathcal{B}^c_m})\Bigr]
	\\
	& \geq \mathbb{E} \Bigl[ 1-\mathbb{P}_{\theta \sim \hat{\rho}_{\lambda}} (\theta \notin\mathcal{A}_m) - \mathbf{1}_{\mathcal{B}^c_m}
	\Bigr]
	\\
	& \geq 1-\delta_m
	\end{align*}
	and thus from \eqref{eq_1_contraction},
	\begin{equation*}
	\mathbb{E} \Bigl[ \mathbb{P}_{\theta \sim \hat{\rho}_{\lambda}} 
	(\theta \in \Theta_m) \Bigr] \geq 1-\delta_m
	.
	\end{equation*}
	This completes the proof.
	
\end{proof}

\subsection{Lemmas}

We begin by presenting a variant of Bernstein’s inequality—Lemma \ref{lemmemassart}—as stated in \cite{MR2319879} (inequality (2.21), Proposition 2.9, p. 24). Following this, Lemma \ref{lemma:dv} introduces the Donsker-Varadhan variational representation, a fundamental result from \cite{catonibook} that plays a central role in the derivation of PAC-Bayesian bounds. And in addition, Lemma \ref{lema_bound_prior_arnak} provides concentration properties of the prior, which are crucial for deriving explicit bounds.

\begin{lemma}
	\label{lemmemassart} Let $U_{1}$, \ldots, $U_{m}$ be independent real
	valued random variables. Suppose there exist two constants
	$v$ and $w$ such that
 for all integers $k\geq 3$,
	$$
	\sum_{j=1}^{m} \mathbb{E}\left[(U_{i})^{k}\right] \leq v\frac{k!w^{k-2}}{2}. 
	$$
	Then, for any $\zeta\in (0,1/w)$,
	$$ 
    \mathbb{E}
\exp\left[\zeta\sum_{j=1}^{m}\left[U_{i}-\mathbb{E}(U_{i})\right]
	\right]
	\leq \exp\left(\frac{v\zeta^{2}}{2(1-w\zeta)} \right) .
    $$
\end{lemma}

\begin{lemma}
	\label{lemma:dv}
Let $h:\Theta\rightarrow\mathbb{R}$ be any function that is both measurable and bounded. Then, the following equation holds:
	\begin{equation*}
	\log \mathbb{E}_{\theta\sim\pi}\left[{\rm e}^{h(\theta)} \right] =\sup_{\rho\in\mathcal{P}(\Theta)}\Bigl[\mathbb{E}_{\theta\sim\rho}[h(\theta)] -	\mathcal{K} (\rho , \pi)\Bigr].
	\end{equation*}
   Furthermore, the supremum over \( \rho \) on the right-hand side is attained by the Gibbs measure \( \pi_h \), whose density is given with respect to \( \pi \).  
	\begin{equation*}
	\frac{{\rm d}\pi_{h}}{{\rm d}\pi}(\theta) =  \frac{{\rm e}^{h(\theta)}}
	{ \mathbb{E}_{\vartheta\sim\pi}\left[{\rm e}^{h(\vartheta)} \right] }.
	\end{equation*}
\end{lemma}

We define the following distribution as a translation of the prior $ \pi $,
\begin{equation}
\label{eq_specific_distribution}
\pi_0(\theta) 
\propto 
\pi (\theta - \theta^*)\mathbf{1}_{
	\{  \| \theta - \theta^*\|_2 \leq 2p \varsigma \} 
}
.
\end{equation}

\begin{lemma}
	\label{lema_bound_prior_arnak}
	Let $p_0 $ be the probability measure defined by (\ref{eq_specific_distribution}). If
	$d\geq 2$ then
	$$
	\int_\Omega \| \theta- \theta^* \|^2 \pi_0({\rm d} \theta)
	\leq
	4p \varsigma^2 
	,
	$$
	and
	$$
	\mathcal{K}(\pi_0 , \pi)
	\leq
	4 s^* \log \left(\frac{H_1 }{\varsigma s^*}\right)
	+
	\log(2)
	.
	$$
\end{lemma}
\begin{proof}
	The proof can be found in \cite{dalalyan2012mirror}.
\end{proof}

\begin{lemma}
	\label{lemma:exponential}
	Under Assumptions~\ref{assume_X_bounded} and~\ref{assum_heavy_tailed}, with $ C_1 = 8 ( \sigma^2 + C^2 ),
	 C_2 = 2^6 \max(\xi,C) C $, put
	\begin{equation}
	\label{defalpha}
	\alpha = \lambda
	-\frac{\lambda^{2} C_1 }{2m(1-\frac{ C_2 \lambda}{m})}
	; \quad
	\beta = \lambda
	+\frac{\lambda^{2} C_1}{2m(1-\frac{ C_2 \lambda}{m})} .
	\end{equation}
	Then for any $\delta\in(0,1)$, and $\lambda \in (0,m/C_2)$,
	\begin{align}
	\mathbb{E} \int \exp  \Biggl\{  \alpha    \Bigl( R(\theta) - R(\theta^*) \Bigr)
	- \lambda ( r(\theta) - r(\theta^*)  )    
	- \log \left[\frac{{\rm d} \hat{\rho}_{\lambda}}{d \pi} (\theta)  \right]
	- \log\frac{2}{\delta}
	\Biggr\}
	\hat{\rho}_{\lambda}(d \theta )
	\leq \frac{\delta}{2}
	\label{lemma:exponential:1}
	\end{align}
	and
	\begin{align}
	\mathbb{E} \sup_{\rho \in \mathcal{P}(\mathbb{R}^{p}) } \exp\Biggl[  \beta
	\left(-\int R{\rm d}\rho + R(\theta^*) \right)
	+ \lambda \left( \int r {\rm d}\rho - r(\theta^*) \right) 
	-
	\mathcal{K}(\rho, \pi) - \log \frac{2}{\delta}\Biggr] \leq
	\frac{\delta}{2}.
	\label{lemma:exponential:2}
	\end{align}
\end{lemma}

\begin{proof}[\textbf{Proof of Lemma~\ref{lemma:exponential}}]
	We start by proving inequality \eqref{lemma:exponential:1}.
	Fix any $ \theta $ with $ | \A_j^\top \theta | \leq C $ and put
	$$ 
	T_{j} =   \left( y_j - (\A_j^\top \theta^* )^2 \right)^{2}
	- \left( y_j - (\A_j^\top \theta )^2 \right)^{2} .
	$$
	Note that $T_1,\dots,T_m $ are independent by construction. We have
	\begin{align*}
	\sum_{j=1}^{m} \mathbb{E}[T_{j}^{2}]  
	&  = 
	\sum_{j=1}^{m} \mathbb{E}
	\left\{
	\left( 2y_j - (\A_j^\top \theta^* )^2 -  (\A_j^\top \theta )^2 \right)^{2}
	\left( (\A_j^\top \theta^* )^2 -  (\A_j^\top \theta )^2 \right)^2
\right\}
	\\
	&  = 
	\sum_{j=1}^{m} \mathbb{E} 
	\left\{ 
	\left[ 2 \varepsilon_j +  (\A_j^\top \theta^* )^2  -  (\A_j^\top \theta )^2  \right]^{2}
	\left[  (\A_j^\top \theta^* )^2 -  (\A_j^\top \theta )^2 \right]^2
\right\}
	\\
	&    \leq
	\sum_{j=1}^{m} \mathbb{E} \left\{ 
	8 ( \varepsilon_j^{2} + C^2) 
	\left[ (\A_j^\top \theta^* )^2 -   (\A_j^\top \theta )^2 \right]^2
	\right\}
	\\
	&    \leq
	\sum_{j=1}^{m}  8 (\sigma^2  + C^2 )
	\mathbb{E} \left[ (\A_j^\top \theta^* )^2 - (\A_j^\top \theta )^2 \right]^2
	\\
	&   \leq 8 m( \sigma^2 + C^2 )
	\left[ R(\theta) - R(\theta^*)\right]
	\\
	& = m
	 C_1\left[ R(\theta) - R(\theta^*)\right]  =:v(\theta, \theta^*).
	\end{align*}
	Next we have, for any integer $k\geq 3$, that
	\begin{align*}
	\sum_{j=1}^{m} \mathbb{E}\left[(T_j )^{k}\right] \leq 
	&
	\sum_{j=1}^{m} \mathbb{E} \left[
	\left| 2y_j - (\A_j^\top \theta^* )^2 -  (\A_j^\top \theta )^2 \right|^{k}
	\left| (\A_j^\top \theta^* )^2 -  (\A_j^\top \theta )^2 \right|^k
	\right]
	\\
	\leq & \sum_{j=1}^{m} \mathbb{E} \left[
	2^{k-1} \left[  |2 \varepsilon_j|^{k} + (2 C)^k  \right]
	\left| (\A_j^\top \theta^* )^2 -  (\A_j^\top \theta )^2 \right|^{k}
	\right]
	\\
	\leq  &  \sum_{j=1}^{m} \mathbb{E} \left[
	2^{2k-1}\left(  |\varepsilon_j|^{k} + C^k \right)
	(2C)^{k-2}
	\left| (\A_j^\top \theta^* )^2 -  (\A_j^\top \theta )^2 \right|^{2}
	\right]
	\\
	\leq    &      2^{2k-1} \left( \sigma^{2}k!\xi^{k-2}
	+ C^k   \right)  (2C)^{k-2}
	\sum_{j=1}^{m}\mathbb{E}  \left| (\A_j^\top \theta^* )^2 - (\A_j^\top \theta )^2 \right|^{2}
	\\
	\leq  &  \frac{ 2^{3k-3} ( \sigma^{2}k!\xi^{k-2} + C^k ) C^{k-2} }{8(\sigma^2 + C^2) } v(\theta, \theta^*)
	\\
	\leq  &  \frac{ 2^{3k-6}  ( \sigma^{2}\xi^{k-2} + C^k ) C^{k-2}}{(\sigma^2 + C^2) } k! v(\theta, \theta^*)
	\\
	\leq  &  2^{3k-5} ( \xi^{k-2} + C^{k-2} ) C^{k-2} k! v(\theta, \theta^*)
	\\
	\leq  &  2^{3k-4} \max(\xi,C)^{k-2} C^{k-2} k! v(\theta, \theta^*)
	\\
	=  &  [2^3 \max(\xi,C) C ]^{k-2} 2^2 k! v(\theta, \theta^*)
	\end{align*}
	and use the fact that, for any $k\geq 3$, $2^2 \leq 2^{3(k-2)}/2 $ to obtain
	\begin{align*}
	\sum_{j=1}^{m} \mathbb{E}\left[(T_{i})^{k}\right] 
	\leq   
	\frac{[ 2^6 \max(\xi,C) C ]^{k-2} k! v(\theta, \theta^*)}{2}
	= 
	v(M, \theta^*) \frac{k!C_2^{k-2}}{2}.
	\end{align*}
	Thus, we can apply Lemma~\ref{lemmemassart} with $U_i := T_i$, $v := v(M, \theta^*)$, $w:=C_2$ and $\zeta := \lambda/m $. We obtain, for any $\lambda\in(0, m/w) = (0, m/C_2)$,
	\begin{align*}
	\mathbb{E} \exp\left[\lambda
	\Bigl( R(\theta)-R(\theta^*)-r(\theta)+r(\theta^*)\Bigr)\right]
& \leq
	\exp\left[\frac{v\lambda^{2}}{2m^{2}(1-\frac{w\lambda}{m})}\right]
\\
& =
	\exp\left[\frac{ C_1
		\left[ R(\theta) - R(\theta^*)\right] \lambda^{2}}{2m(1-\frac{C_2 \lambda}{m})}\right].
	\end{align*}
	Rearranging terms, and using the definition of $\alpha$ (that is~\eqref{defalpha}),
	$$
	\mathbb{E} \exp\left[\alpha
	\Bigl( R(\theta) - R(\theta^*) \Bigr)
	- \lambda ( r(\theta) - r(\theta^*) ) \right] \leq 1.
	$$
	Multiplying both sides by $ \delta/2$ and then integrating w.r.t. the probability distribution $ \pi(.) $, we get
	$$
	\int \mathbb{E} \exp\Biggl[\alpha
	\Bigl( R(\theta) - R(\theta^*) \Bigr)
	- \lambda ( r(\theta) - r(\theta^*)  )
	- \log\frac{2}{\delta}\Biggr]  \pi (d \theta) \leq \frac{\delta}{2}.
	$$
	Next, Fubini's theorem gives
	$$
	\mathbb{E} \int  \exp\Biggl[\alpha
	\Bigl( R(\theta) - R(\theta^*) \Bigr)
	- \lambda ( r(\theta) - r(\theta^*) )
	- \log\frac{2}{\delta}\Biggr]  \pi (d \theta) \leq \frac{\delta}{2}.
	$$
	and note that for any measurable function $h$,
	\begin{align*}
	\int  \exp[h(\theta)]  \pi (d \theta)
	= 
	\int  \frac{\exp[h(\theta)] } { \frac{{\rm d} \hat{\rho}_{\lambda}}{{\rm d}\pi}(\theta) } \hat{\rho}_{\lambda}(d \theta)
	= 
	\int  \exp\left[h(\theta)- \log \frac{{\rm d} \hat{\rho}_{\lambda}}{{\rm d}\pi}(\theta) \right] \hat{\rho}_{\lambda}(d \theta)
	\end{align*}
	to get \eqref{lemma:exponential:1}.
	
	Let us now prove \eqref{lemma:exponential:2}. Here again, we start with an application of Lemma~\ref{lemmemassart}, but this time with $U_i := - T_i$ (we keep $v := v(M, \theta^*)$, $w:=C_2$ and $\zeta := \lambda/m $). We obtain, for any $\lambda\in(0, m/C_2)$,
	$$
	\mathbb{E}  \exp\left[\lambda
	\Bigl( r(\theta) - r(\theta^*) - R(\theta) + R(\theta^*)\Bigr)\right]
	\leq
	\exp\left[\frac{ C_1
		\left[ R(\theta) - R(\theta^*)\right] \lambda^{2}}{2m(1-\frac{C_2 \lambda}{m})}\right].
	$$
	Rearranging terms, using the definition of $\beta$ (that is~\eqref{defalpha}) and multiplying both sides by $ \delta/2 $, we obtain
	\begin{equation*}
	\mathbb{E} \exp\Biggl[\beta
	\left(-R(\theta) + R(\theta^*) \right)
	+ \lambda \left( r(\theta)- r(\theta^*) \right) - \log \frac{2}{\delta}\Biggr] \leq
	\frac{\delta}{2}.
	\end{equation*}
	We integrate with respect to $\pi$ and use Fubini's theorem to get:
	\begin{equation*}
	\mathbb{E} \int \exp\Biggl[\beta
	\left(-R(\theta) + R(\theta^*) \right)
	+ \lambda \left( r(\theta)- r(\theta^*) \right) - \log \frac{2}{\delta}\Biggr] \pi({\rm d} \theta ) \leq
	\frac{\delta}{2}.
	\end{equation*}
	Here, we use a different argument from the proof of the first inequality: we use Lemma \ref{lemma:dv} on the integral, this gives directly \eqref{lemma:exponential:2}.
\end{proof}

\end{document}